\documentclass[letterpaper]{article} %
\usepackage{aaai23}  %
\usepackage{times}  %
\usepackage{helvet}  %
\usepackage{courier}  %
\usepackage[hyphens]{url}  %
\usepackage{graphicx} %
\usepackage{natbib}  %
\usepackage{caption} %
\title{Common Knowledge of Abstract Groups}
\author {
    Merlin Humml,
    Lutz Schröder
}
\affiliations {
  Friedrich-Alexander-Universität Erlangen-Nürnberg, Erlangen, Germany\\
  \{merlin.humml, lutz.schroeder\}@fau.de
}

\usepackage[nomargin,marginclue,footnote,draft]{fixme}
\FXRegisterAuthor{ls}{als}{LS}
\FXRegisterAuthor{mg}{amg}{MG}

\usepackage{breqn}
\usepackage{amsmath, amssymb, amsthm, stmaryrd}

\usepackage{acronym}
\usepackage{xspace}
\usepackage{csquotes}
\usepackage[minimal]{yhmath}
\renewcommand{\hat}{\widehat}
\usepackage[conf={normal, end/.append style={restate}, no link to proof, one big link={}}]{proof-at-the-end}

\newcommand{\ExpTime}{\mbox{\textsc{ExpTime}}\xspace}
\newcommand{\ALC}{\ensuremath{\mathcal{ALC}}\xspace}
\newcommand{\ALCO}{\ensuremath{\mathcal{ALCO}}\xspace}
\newcommand{\Ind}{\mathit{Ind}}
\newcommand{\nec}{\mathit{Nec}}
\newcommand{\anti}{\mathit{AM}}

\newcommand{\AProps}{\mathsf{AP}}
\newcommand{\Prog}{\mathsf{Prog}}
\newcommand{\var}{z}
\newcommand{\Land}{\bigwedge}

\newcommand{\progbox}[1]{[#1]}
\newcommand{\progdiamond}[1]{\langle #1\rangle}

\newcommand{\defeq}{\ensuremath{:=}}
\newcommand{\Sem}[1]{\ensuremath{\llbracket #1 \rrbracket}}
\newcommand{\Atoms}{\ensuremath{\mathsf{At}_{\mathsf{W}}}\xspace}
\newcommand{\AAtoms}{\ensuremath{\mathsf{At}_\Ag}\xspace}
\newcommand{\Ag}{\mathsf{Ag}}
\newcommand{\Wl}{\mathsf{W}}
\newcommand{\atom}{p}

\newcommand{\Agents}{\ensuremath{Ag}\xspace}
\newcommand{\agent}{\ensuremath{a}}

\newcommand{\Worlds}{\ensuremath{X}\xspace}
\newcommand{\world}{\ensuremath{x}\xspace}
\newcommand{\oworld}{\ensuremath{y}\xspace}
\newcommand{\Pow}{\ensuremath{\mathcal{P}}}

\newcommand{\M}{\ensuremath{\mathcal{M}}}

\newcommand{\C}{\ensuremath{\mathcal{C}}}

\newcommand{\A}{\ensuremath{\mathcal{A}}}
\newcommand{\LA}[0]{\relax\ifmmode {\mathcal{L}_{\mathsf{Ag}}} \else {agent logic}\fi}
\newcommand{\SemM}[2][\M]{\relax\ifmmode {\Sem{#2}_{#1}} \else {extension of \ensuremath{#2} in \ensuremath{#1}}\fi}

\newcommand{\SemMA}[2][\A]{\relax\ifmmode {\Sem{#2}_{#1}} \else {agent-extension of \ensuremath{#2} in \ensuremath{#1}}\fi}
\newcommand{\MAt}[2][\M]{\ensuremath{#1,#2}}
\newcommand{\SatAt}[3][\M]{\relax\ifmmode {\MAt[#1]{#2} \models #3}\else {#2{} satisfies~\ensuremath{#3}}\fi}
\newcommand{\SatAtA}[3][\A]{\relax\ifmmode {\MAt[#1]{#2} \models #3}\else {#1{} satisfies~\ensuremath{#3} at \ensuremath{#2}}\fi}
\newcommand{\nSatAt}[3][\M]{\relax\ifmmode {\MAt[#1]{#2} \not\models #3}\else {#1{} does not satisfy \ensuremath{#3} at \ensuremath{#2}}\fi}
\newcommand{\nSatAtA}[3][\A]{\relax\ifmmode {\MAt[#1]{#2} \not\models #3}\else {#1{} does not satisfie \ensuremath{#3} at \ensuremath{#2}}\fi}

\renewcommand{\implies}{\rightarrow}
\renewcommand{\iff}{\leftrightarrow}

\acrodef{CIEL}[AGEL]{abstract-group epistemic logic}
\acrodef{GEL}{group epistemic logic}

\acrodef{PDL}{propositional dynamic logic}

\theoremstyle{definition}
       \newtheorem{definition}{Definition}
       \newtheorem{example}[definition]{Example}
       \newtheorem{remark}[definition]{Remark}
\theoremstyle{plain}

       \newtheorem{corollary}[definition]{Corollary}

\begin{document}
\setquotestyle{british}
\maketitle

\begin{abstract}
  Epistemic logics typically talk about knowledge of individual agents
  or groups of explicitly listed agents. Often, however, one wishes to
  express knowledge of groups of agents specified by a given property,
  as in \textquote{it is common knowledge among economists}. We
  introduce such a logic of common knowledge, which we term
  \emph{\ac{CIEL}}. That is, \ac{CIEL} features a common knowledge
  operator for groups of agents given by concepts in a separate agent
  logic that we keep generic, with one possible agent logic
  being~$\ALC$. We show that \ac{CIEL} is \ExpTime-complete, with the
  lower bound established by reduction from standard group epistemic
  logic, and the upper bound by a satisfiability-preserving embedding
  into the full \(\mu\)-calculus. Further main results include a
  finite model property (not enjoyed by the full \(\mu\)-calculus) and
  a complete axiomatization.
\end{abstract}

\acresetall{} %

\section{Introduction}\label{sec:introduction}

Epistemic (modal) logic is concerned with the individual and
collective knowledge of agents. One of the most important modalities
for collective knowledge is \emph{common knowledge}: A fact~$\phi$ is
common knowledge in a given group of agents if everyone in the group
knows~$\phi$, and everyone knows that everyone knows~$\phi$, etc. In
the present work, our focus of attention is on the involved notion of
group of agents. The most basic variant of the common knowledge
operator, typically written~$C$, refers to \emph{all} agents in a
predetermined finite set~$\Agents$ that forms a parameter of the logic
as a whole \cite{DBLP:books/mit/FHMV1995}. In a more fine-grained
variant,~$C$ can be annotated with an explicitly given subset of the
set of agents: For $A\subseteq\Agents$, $C_A\phi$ says that~$\phi$ is
common knowledge among the agents in~$A$. For instance, if
$\mathsf{Alice}$ and $\mathsf{Bob}$ are legitimate participants in a
communication protocol and~$\phi$ is a fact concerning a shared key,
then~$\phi$ would ideally be common knowledge of $\mathsf{Alice}$ and
$\mathsf{Bob}$ but not of a malicious third party $\mathsf{Charlie}$
-- i.e. $C_{\{\mathsf{Alice},\mathsf{Bob}\}}\phi$ would hold but
$C_{\{\mathsf{Alice},\mathsf{Bob},\mathsf{Charlie}\}}\phi$ would not.

Listing agents in a group explicitly is appropriate in well-controlled
settings such as the above, where the participants in the epistemic
situation are fixed and previously known. In other application
contexts, however, this may not always be the case, in particular in
statements found in real-world argumentation.  Consider, for example,
the sentence \textquote{Doctors agree that smoking is bad for your
  health.}  We take this sentence (maybe debatably) as making a
statement about common knowledge of all doctors. Encoding this claim
as a formula of the form $C_A\phi$ where~$A$ is a finite set
explicitly enumerating all doctors is clearly neither feasible nor
even semantically desirable, as the statement is presumably meant to
hold without regard to exactly how many, and which, doctors are
practising in the world at the moment. Rather, one would want~$A$ to
be given by the defining property of \emph{being a doctor}.

In the present paper, we introduce an epistemic logic that allows
precisely this: \emph{\ac{CIEL}} features a common knowledge operator
for groups of agents described by concepts in a dedicated \emph{agent
  logic}. We keep the technical development generic in the choice of
agent logic, subject to some technical requirements on the agent logic
that are satisfied, for instance, by the description logic~$\ALC$; so
we can describe groups of agents such as \textquote{doctors and
  pharmacists} or \textquote{parents of teenagers}. We note that we
treat the agent logic as rigid, i.e.\ there is no uncertainty about
membership in the groups it describes. In other words, group
descriptions are \emph{de re} rather than \emph{de dicto}. We further
illustrate the logic on a variant of the muddy children puzzle where
the number of participants is unspecified and potentially large. Our
main results on \ac{CIEL} are \ExpTime-completeness of the
satisfiability problem; a bounded (specifically, doubly exponential)
model property; and a complete axiomatization. Technically, we
establish the lower complexity bound by a satisfiability-preserving
translation of standard group epistemic logic
\cite{DBLP:books/mit/FHMV1995} into \ac{CIEL}, and the upper bound by
a satisfiability-preserving translation of \ac{CIEL} into the full
\(\mu\)-calculus (i.e.~$\ALC$ with inverse roles and fixpoint
operators~\cite{Vardi98}). Use of the full \(\mu\)-calculus avoids the
exponential blow-up that would be incurred by a more naive reverse
translation of \ac{CIEL} into group epistemic logic. However, the full
$\mu$-calculus does not have the finite model
property~\cite{Vardi98,Streett82}. Instead, we show the bounded model
property and completeness by means of a filtered model construction
that uses ideas from the finite model construction for propositional
dynamic logic~\cite{FischerLadner79,blackburn_rijke_venema_2001}, in
particular transitive closure of (small) canonical pseudo-models.

\subsubsection*{Related Work}\label{sec:related-work}

There is a line of research on indexing knowledge modalities with
\emph{names} that designate groups of
agents~\cite{DBLP:journals/logcom/GroveH93} \cite[Chapter
6]{DBLP:books/mit/FHMV1995}. We refer to such groups as \emph{named
  groups}; they are similar to the atoms of our agent logic but are
\emph{non-rigid}, i.e.~their interpretation depends on the current
world, in an approach that is focused on the analysis of knowledge
about the identity of agents. Although common knowledge of such
name-defined groups has been mentioned early
on~\cite{DBLP:journals/logcom/GroveH93}, results have largely focused
on operators of the type \textquote{every agent / some agent with name~$n$
knows} (recall that generally, \textquote{everyone knows~$\phi$} differs
from~$\phi$ being common knowledge in that it need not imply that
everyone knows that everyone knows~$\phi$, etc.). Recently,
\citet{DBLP:journals/corr/abs-2106-11493} have shown
completeness and the finite model property (but no complexity bound)
for common knowledge of name-defined groups. (A different form of
common knowledge for non-rigid groups has been considered earlier,
without considering
axiomatization~\cite{MosesTuttle88}.)

Extending the descriptive means for groups of agents has been
considered already by \citet{DBLP:journals/logcom/GroveH93}, who give an
axiomatization (but no complexity bound) of a logic that features
\textquote{everyone knows} and \textquote{someone knows} operators for propositional
combinations of names. Additional expressive means are provided by
subsequent first-order extensions of the
logic %
that allow quantifying over agent
names~\cite{DBLP:journals/ai/Grovez95,DBLP:journals/rsl/NaumovT19} in
the style of term modal
logic~\cite{DBLP:journals/sLogica/FittingTV01}. Quantifying over
agents provides a straightforward way of encoding, e.g., the
\textquote{everybody knows operator}; for instance, \textquote{every doctor knows~$\phi$}
would be expressed as $\forall x.\,\mathsf{doctor}(x)\implies K_x\phi$
if~$K_x$ denotes the usual single-agent knowledge modality \textquote{$x$
knows}. Common knowledge, on the other hand, involves transitive
closure and as such is not first-order expressible, hence not easily
accommodated in such frameworks. Also, decidability results in a
first-order setting will, of course, require additional restrictions
(e.g.~it has recently been shown that the two-variable fragment of
term-modal logic is decidable, with complexity between
\textsc{NExpTime} and
\textsc{2ExpSpace}~\cite{PadmanabhaRamanujam19}).

The above logics and ours employ ternary transition relations relating
agents to pairs of worlds. Beyond epistemic logic, ternary relations
appear, e.g., in the logic of Pierce algebras~\cite{derijkephd}, in
arrow logic \cite{venemaCrashArrow}, and in Routley-Meyer-style
semantics of \emph{relevance logic}~\cite{sep-logic-relevance}. While
such modalities are formally similar to \emph{everyone-knows}
modalities for abstract groups, the frame conditions imposed on models
and, consequently, the attached meta-theory are quite different from
ours.

\section{Abstract-Group Epistemic Logic}\label{sec:conc-index-epist}

\noindent We proceed to introduce the syntax and semantics of
\emph{\acf{CIEL}}.

\paragraph*{Syntax} We parametrize the logic over the choice of an
\emph{agent logic}~\(\LA\) that serves to specify groups of agents. We
will discuss assumptions on the semantics of \(\LA\) later in this
section. Syntactically, we require only that \(\LA\) has a formula
syntax where formulae are expressions in some grammar, in particular
giving rise to a standard notion of \emph{subformula}, and includes
propositional atoms from a set~$\AAtoms$, referred to as \emph{agent
  atoms}, as well as the full set of Boolean connectives. Properties
of~\(\LA\) needed in our main results will be named explicitly in the
respective theorems. One choice for \(\LA\) that satisfies all
requisite properties is the standard description logic
$\ALC$~\cite{BaaderEA03}.

We further assume a set \Atoms of \emph{world atoms}. The set of
\emph{(world) formulae} $\phi,\chi,\dots$ of \ac{CIEL} is then defined
by the grammar
\begin{equation*}
  \phi,\chi \defeq \bot \mid \atom \mid \neg \phi \mid \phi \land \chi
  \mid C_\psi \phi \qquad(\atom \in \Atoms, \psi \in
  \LA).
\end{equation*}
\noindent That is, we include propositional atoms and Boolean
connectives; further Boolean connectives $\lor$, $\implies$, $\iff$
are defined as usual. The key feature is then the common knowledge
operator $C_\psi$ for groups of agents defined by an agent
formula~$\psi$; a formula \(C_\psi \phi\) is read \textquote{\(\phi\) is common
knowledge among agents satisfying \(\psi\)}, using the term
\emph{common knowledge} in the sense recalled in the introduction.
\begin{example}
  We may encode the statement \textquote{parents of teenagers know
    that education is pointless} as the \ac{CIEL} formula
  $C_{\exists\mathsf{hasChild}.\,\mathsf{Teenager}}\, \mathsf{ep}$,
  using $\ALC$ as the agent logic and understanding the world atom
  $\mathsf{ep}$ as \textquote{education is pointless}. The even more
  frustrating fact that parents know that their offspring know this as
  well would be captured by the formula
  $C_{\exists\mathsf{hasChild}.\,\mathsf{Teenager}}\,C_{\mathsf{Teenager}}\,
  \mathsf{ep}$.
\end{example}

\paragraph*{Semantics} We assume that the agent logic comes with a
notion of \emph{agent model}, and that every agent model $\A$ is
equipped with an underlying set $\Agents$ of \emph{agents} and a
satisfaction relation $\models_\A\:\subseteq\Agents\times\LA$; we
write $\A,a\models\psi$ for $(a,\psi)\in\;\models_\A$, and
$\Sem{\phi}_\A=\{a\in\Agents\mid \A,a\models \psi\}$. We require that
\emph{\(\LA\) conservatively extends classical propositional
  logic}. By this we mean more specifically that \(\LA\) does not
impose restrictions on valuations of agent atoms, i.e.\ given a set
$\Agents$ and a valuation $V_\Ag\colon\AAtoms\to\Pow(\Agents)$, there
always exists an agent model~$\A$ with underlying set~$\Agents$ such
that $\A,a\models q$ iff $a\in V_\Ag(q)$, for $a\in\Agents$ and
$q\in\AAtoms$.

Then, an \emph{(\ac{CIEL}) model} \(\M = (\Worlds, \A, V_\Wl,\sim)\)
consists of a set \Worlds of \emph{worlds}, an agent model~\(\A\), a
\emph{world valuation}~\(V_\Wl\colon\Atoms\to\Pow(\Worlds)\)
interpreting the world atoms, and a family~$\sim$ of
\emph{indistinguishability relations}
\(\sim_a\;\subseteq\Worlds\times\Worlds\) indexed over
agents~$a\in\Agents$. We require that each~$\sim_a$ is an equivalence
relation (in keeping with the usual view that epistemic
indistinguishability relations should be equivalence relations); see
however Remark~\ref{rem:s5}. For a set $A\subseteq\Agents$ of agents, we
write $\sim_A=(\bigcup_{a\in A}\sim_a)^*$ \footnote{This should not be
  confused with similar notation used to denote the intersection of
  the~$\sim_a$ in work on epistemic logic with quantification over agents~\cite{DBLP:journals/rsl/NaumovT19}} where $(-)^*$ denotes reflexive-transitive closure
(note that $\sim_A$ is symmetric, hence an equivalence). We define
satisfaction $\SatAt{x}{\psi}$ (\emph{\SatAt{x}{\psi}}) of a
formula~$\psi$ at a world~$x$ recursively by
  \begin{gather*}
    \nSatAt{\world}{\bot}  \qquad \SatAt{\world}{\atom} \text{ iff } \world \in V_\Wl(\atom)\\
    \SatAt{\world}{\neg\phi} \text{ iff } \nSatAt{\world}{\phi}\\
    \SatAt{\world}{\phi \land \chi} \text{ iff } \SatAt{\world}{\phi} \text{ and } \SatAt{\world}{\chi}\\
    \SatAt{\world}{C_\psi\phi} \text{ iff whenever }\world
    \sim_{\SemMA{\psi}} \oworld,\text{ then } \SatAt{\oworld}{\phi};
  \end{gather*}
  that is, $C_\psi$ is the standard box modality for
  $\sim_{\SemMA{\psi}}$.  When $\SatAt{\world}{\phi}$, then we also
  say that $(\M,\world)$ is a \emph{model of~$\phi$} (and we will use
  this phrasing in general, also for other logics). The formula
  \(\phi\) is \emph{satisfiable} if there is a model of~\(\phi\), and
  \emph{valid} (notation: \(\models\phi\)) if \(\M,\world\models\phi\)
  for all \(\M,\world\). We write
  \(\SemM{\phi}=\{\world\in\Worlds\mid\SatAt{\world}{\phi}\}\).

  We record a fixpoint characterization of $C_\psi$:
\begin{theoremEnd}[end]{lemma}\label{thm:C_trans}
  The set \(\SemM{C_\psi\phi}\) is the greatest fixed point of the
  function $F\colon\Pow(\Worlds)\to\Pow(\Worlds)$ given by $F(U)$
  being the set of worlds~$x$ such that $\SatAt{\world}{\phi}$ and
  whenever $\SatAtA{\agent}{\psi}$ and $\world \sim_{\agent} \oworld$,
  then $\oworld \in U$.
\end{theoremEnd}
\begin{proofEnd}
  By the Knaster-Tarski fixpoint theorem, we can equivalently show
  that \(\SemM{C_\psi\phi}\) is the greatest postfixed point
  of~$F$. First, it is clear that \(\SemM{C_\psi\phi}\) is a postfixed
  point of~$F$: Since~$C_\psi$ is the box operator for the transitive
  and reflexive closure of the $\psi$-successor relation, every
  world~$\world$ in \(\SemM{C_\psi\phi}\) satisfies~$\phi$, and all
  $\psi$-successors of~$\world$ satisfy $C_\psi\phi$ again. Second,
  if~$K\subseteq\Worlds$ is another postfixed point of~$F$,
  i.e.~$K\subseteq F(K)$, then every world in~$K$ satisfies~$\phi$,
  and~$K$ is closed under $\psi$-successors, so by induction on the
  formation of~$\sim_\psi$ as a reflexive-transitive closure, every
  world in~$K$ is in \(\SemM{C_\psi\phi}\).
\end{proofEnd}

\begin{remark}\label{rem:s5}
  Since the modality $C_\psi$ takes reflexive-transitive closures, it
  is in fact immaterial whether the indistinguishability
  relations~$\sim_a$ are individually reflexive and transitive; we
  impose the corresponding requirement mainly to ease notation and
  discussion.
\end{remark}

\begin{theoremEnd}[end, text proof={Details for \string\autoref{thm:prAtEnd\pratendcountercurrent}}]{remark}\label{rem:agent-knowledge}
  In logical settings with non-rigid (i.e.~world-dependent) agent
  models
  (e.g.~\cite{DBLP:journals/logcom/GroveH93,DBLP:journals/ai/Grovez95,DBLP:journals/rsl/NaumovT19}),
  one can accommodate uncertainty about the identity and properties of
  agents, and moreover specify agent groups by their knowledge, as in
  `people who know~$\phi$ also know~$\psi$'. As indicated in the
  introduction, \ac{CIEL}, which assumes the agent logic to be rigid,
  trades this mode of expression for a computationally and
  axiomatically tractable treatment of common knowledge. %
  One may still envisage extending \ac{CIEL} with an operator~$I$, with
  $I\,\phi$ read as \textquote{knows about the truth or falsity
    of~$\phi$}. That is, an agent satisfies $I\,\phi$ if in every
  world, she knows either~$\phi$
  or~$\neg\phi$. %
  For instance, if~$\phi$ represents the publicly contested fact
  whether or not transfer negotiations are under way concerning a
  football player~$P$, then $I\,\phi$ would hold for the relevant
  officials of the involved clubs and maybe for~$P$. However,
  harnessing such an operator technically is likely to be
  challenging. To name one potential obstacle, the first-order
  translation of~$I\,\phi$ would presumably need to involve three
  variables (i.e.~unlike the translation of most modal logics would
  not end up in the two-variable fragment): one for the agent~$x$, and
  two for worlds that are indistinguishable for~$x$, and then required
  to agree on~$\phi$ (see the
  appendix). %
\end{theoremEnd}
\begin{proofEnd}
  Using  notation for models employed in the definition of the semantics
  directly in first-order syntax, we translate satisfaction of the
  formula~$I\,\phi$ by an agent~$\agent$ roughly into
  \begin{equation*}
    \forall \world,\oworld.\,\world\sim_\agent \oworld\to (\bar\phi(\world)\iff \bar\phi(\oworld))
  \end{equation*}
  where~$\bar\phi(-)$ represents a first-order translation of~$\phi$.
  This formula needs three variables $a,x,y$. (This observation is
  purely heuristic; we do not formally claim that use of three variables
  cannot be avoided, nor would a two-variable translation directly imply
  decidability of the extension of \ac{CIEL} with~$I$, as models are
  subject to constraints that fail to be first-order definable.)
\end{proofEnd}

\begin{remark}
  Since \ac{CIEL} is effectively a fixpoint logic, it is expected that
  compactness fails. Indeed, given atomic agent concepts $A,B$ and a
  world atom~$p$, the set consisting of the formula
  $\neg C_{A\lor B}\:p$ and all formulae of the form
  $C_{D_1}C_{D_2}\dots C_{D_n}p$, for $n\ge 0$ and
  $D_1,\dots,D_n\in\{A,B\}$, is unsatisfiable but all its finite
  subsets are satisfiable. It follows that there is no finitary proof
  system for \ac{CIEL} that is \emph{strongly} complete, i.e.~makes
  all unsatisfiable sets of formulae inconsistent.  We later give a proof system that is
  \emph{weakly} complete, i.e.~derives all valid formulae.
\end{remark}

\section{Complexity}

\noindent We show next that the satisfiability problem of \ac{CIEL} is
\ExpTime-complete. 

\subsection{Lower Bound:\\
  Reduction from Group Epistemic Logic}
We prove \ExpTime-hardness by a satisfiability-preserving encoding of
standard \acf{GEL} (with common knowledge), which is known to be
\ExpTime-hard~\cite{DBLP:books/mit/FHMV1995}. (To facilitate the
subsequent discussion, we reduce from a slightly more expressive logic
than strictly necessary for the hardness proof.) We briefly recall the
syntax and semantics of \ac{GEL}: The logic is parametrized over a
finite set $\Agents$ of agents and a set $\Atoms$ of (world)
atoms. The set of \emph{formulae}~$\phi,\psi,\dots$ of \ac{GEL} is
then given by the grammar
\begin{equation*}
  \phi,\psi::= \bot\mid \atom\mid \neg\phi\mid\phi\land\psi\mid C_G\phi
\end{equation*}
where $\atom\in\Atoms$ and $\emptyset\neq G\subseteq\Agents$, with
$C_G\phi$ read \textquote{$\phi$ is common knowledge among the agents in~$G$}.
(Knowledge operators $K_\agent$ for individual agents~$\agent$ are
included as common knowledge operators~$C_{\{\agent\}}$.) As indicated
in the introduction, the difference with \ac{CIEL} is that in \ac{GEL}, groups~$G$
of agents need to be given as enumerated finite subsets of a known
fixed set of named agents.  Models $\M=(\Worlds,V_\Wl,\sim)$ consist of
a set~$\Worlds$ of worlds, a (world) valuation
$V_\Wl\colon\Atoms\to\Pow(\Worlds)$, and a family~$\sim$ of
indistinguishability relations
$\sim_\agent\;\subseteq\Worlds\times\Worlds$, indexed over agents
$\agent\in\Agents$ and again required to be equivalence relations. For
$G\subseteq\Agents$, we write
$\sim_G:=(\bigcup_{\agent\in G}\sim_\agent)^*$. Then, satisfaction
$\SatAt{\world}{\phi}$ of a formula~$\phi$ at a world~$\world$ is
defined recursively by the expected clauses for atoms and
propositional connectives, and
\begin{equation*}
  \SatAt{\world}{C_G\phi}\text{ iff whenever }\world\sim_G\oworld,
  \text{ then }\SatAt{\oworld}{\phi}.
\end{equation*}%
\newcommand{\gctrans}{q}%
The encoding~$\gctrans$ of \ac{GEL} into \ac{CIEL} is given as
follows. We introduce a fresh agent atom \(p_\agent\) for each
\(\agent \in \Agents\). For a \ac{GEL} formula~$\phi$,~$\gctrans(\phi)$ is
then defined recursively by
\begin{equation*}
  \gctrans(C_G \phi) \defeq C_{\bigvee_{\agent \in G} p_\agent} \gctrans(\phi)
\end{equation*}
and commutation with all other constructs
(i.e.~$\gctrans(\neg\phi)=\neg\gctrans(\phi)$ etc.). Using the running
assumption that \(\LA\) conservatively extends classical propositional
logic, we obtain

\begin{theoremEnd}[end]{theorem}[Lower complexity
  bound]\label{thm:ciel-hardness}
  The satisfiability problem for \ac{CIEL} is \ExpTime-hard.
\end{theoremEnd}
\begin{proof}[Proof sketch] We need to show that~$\gctrans$ is indeed
  satisfiability-preserving. A model $(\M,\world)$ of a \ac{GEL}
  formula~$\phi$ over the set $\Agents$ of agents, with
  $\M=(\Worlds,V_\Wl,\sim)$, is transformed into a model
  $(\M',\world)$ of the \ac{CIEL} formula $\gctrans(\phi)$, with
  $\M'=(\Worlds,\A,V_\Wl,\sim')$, by taking $\Agents$ to be the
  underlying set of~$\A$, and $V_\Ag(p_\agent)=\{\agent\}$ for
  $\agent\in\Agents$; this uses the running assumption that \(\LA\)
  conservatively extends classical propositional logic.

  Conversely, a model $(\M,x)$ of $\gctrans(\phi)$, with
  $\M=(\Worlds,\A,V_\Wl,\sim)$, is transformed into a model
  $(\M',\world)$ of~$\phi$, with $\M'=(\Worlds,V_\Wl,\sim')$, by
  taking $\sim'_\agent\;=\;\sim_{V_\Ag(p_\agent)}$ (using  notation like in the formal semantics of \ac{CIEL}).
\end{proof}
\begin{proofEnd}
  By reduction from \ac{GEL} satisfiability. We show that the
  translation $\gctrans$ given in the main body of the paper is
  satisfiability-preserving, i.e.~that~$\phi$ is satisfiable iff
  $\gctrans(\phi)$ is satisfiable.

  \emph{\textquote{If}:} Given an \ac{CIEL}-model
  \(\M = (\Worlds, (\Agents, V_\Ag), V_\Wl,\) \({\sim})\), we construct a
  \ac{GEL}-model \(\M' = (\Worlds, \Agents, V_\Wl,\sim')\) by taking
  \begin{align*}
    \sim'_\agent \;=\;\sim_{\SemM[\A]{p_\agent}}.
  \end{align*}
  We show by induction on \(\phi\) that for all
  \(\world \in \Worlds\), \(\SatAt[\M]{\world}{\gctrans(\phi)}\) iff
  \(\SatAt[\M']{\world}{\phi}\).  The cases for world atoms and
  Boolean operators are trivial; we do the case for \(C_G \phi\). By
  definition,
  $\gctrans(C_G\phi)=C_{\bigvee_{a\in G}p_a}\gctrans(\phi)$. Now
  \(C_{\bigvee_{a\in G}p_a}\) is a standard modal box on the
  transitive-reflexive closure of the union of the relations
  $\sim_{\Sem{p_a}_\A}$ over all~$a\in G$, and by construction of
  $\sim'$, $C_G$ is a standard modal box over the same relation. We
  are done by induction.

  \emph{\textquote{Only if}:} Given a \ac{GEL} model $\M=(\Worlds,V_\Wl,\sim)$
  over the set $\Agents$ of agents, we construct an \ac{CIEL} model
  $\M'=(\Worlds,\A,V_\Wl,\sim')$ by taking $\Agents$ to be the
  underlying set of~$\A$, and $V_\Ag(p_\agent)=\{\agent\}$ for
  $\agent\in\Agents$; this uses the running assumption that \(\LA\)
  conservatively extends classical propositional logic, so an agent
  model~$\A$ with the prescribed valuation of agent atoms is
  guaranteed to exist. The truth conditions in the respective
  semantics of~\ac{GEL} and \ac{CIEL} are then literally the same.
\end{proofEnd}

\newcommand{\cgtrans}{s}
\begin{remark}
  Depending on additional restrictions on the agent logic, it will
  sometimes be possible to give also a fairly straightforward
  satisfiability-preserving translation in the reverse direction, from
  \ac{CIEL} to \ac{GEL}. For instance, if the agent logic is just
  classical propositional logic, then we can proceed as follows:
  Let~$\phi$ be an \ac{CIEL} formula, and let $A\subseteq\AAtoms$ be the
  set of agent atoms mentioned in~$\phi$. Let~$\Agents$ be the
  (finite) set of truth valuations $\kappa\colon A\to 2$ in the set
  $2=\{\bot,\top\}$ of Boolean truth values, and write $\kappa(\psi)$
  for the truth value of a propositional formula~$\psi$ over~$A$
  under~$\kappa$; then a translation of~$\phi$ into a
  satisfiability-equivalent \ac{GEL} formula $\cgtrans(\phi)$ is
  defined recursively by
  \begin{equation*}
    \cgtrans(C_\psi\chi)=C_{\{\kappa\in\Agents\mid\kappa(\psi)=\top\}}\cgtrans(\chi)
  \end{equation*}
  and commutation with all other constructs.  However, even in this
  basic case, such a translation will be of limited use as it has
  exponential blowup (the set
  ${\{\kappa\in\Agents\mid\kappa(\psi)=\top\}}$ can be exponentially
  large). For more expressive agent logics, e.g.\ whenever the agent
  logic extends \ALC, one has (series of) formulae that are
  satisfiable only over exponentially large agent models: Given agent
  formulae~$\psi_n$ that are of polynomial size in~$n$ but satisfiable
  only over agent models of exponential size in~$n$ (in \ALC,
  such~$\psi_n$ exist), the \ac{CIEL} formulae
  $p\land C_{\psi_n}\neg p$ are satisfiable only over models whose
  agent model components are of exponential size in~$n$. Indeed, from
  a purely computational point of view (and for suitably restricted
  agent logics), one may see \ac{CIEL} as a way of dealing with
  exponentially many agents without incurring doubly exponential
  computational cost. We realize this by an encoding into a different
  target logic, discussed after the next example.
\end{remark}

\begin{example}[Lots of muddy children]
  The classical \emph{muddy children} puzzle with \(k\) many children can be seen as \(k\) many agents \(A_i\) communicating according to a fixed protocol  to gain common knowledge of a length-\(k\)  bitstring where each agent \(A_i\) can see all bits except the one at index \(i\) \cite[Section~4.3]{Pavlovic21}. Specifically, it is commonly known initially that at least one bit is set, and the protocol then proceeds in rounds in which the agents announce whether they have learned their missing bit. The full modelling of the puzzle thus requires a dynamic epistemic logic with common knowledge and public announcements~\cite{BaltagEA98,Lutz06}. Here, we concentrate on modelling, in an extended setting,  how knowledge is gained in individual rounds of the protocol, which does not require public announcements; we generalize the  textbook treatment by Huth and Ryan~\shortcite{DBLP:books/daglib/0017977}.
  
  We consider a variant of the puzzle that can essentially be seen as
  a product of~$n$ copies of the original puzzle. We then have an
  \(n \times k\)-matrix of bits, and each agent has an
  \emph{invisibility type} consisting of one \emph{invisibility index}
  per row determining the bit she cannot see in that row (in the
  original puzzle, there is only one row, and the invisibility type is
  the identity of the agent).  We require that every bit of the matrix
  is seen by at least one agent. We do not otherwise restrict which
  invisibility types are realized; also, a given invisibility type may
  be realized by more than one agent. Note that there are
  exponentially many (viz, $k^n$) invisibility types; the point of
  these considerations being that the number of (distinguishable)
  agents is a)~not fixed, and b)~potentially large.  We introduce
  propositional atoms \(p_{(j,i)}\) for \(1 \le j \le n\) and
  \(1 \le i \le k\) indicating whether the bit at position \((j,i\))
  in the matrix is \emph{set}~(\(1\)).  We use agent atoms~\(h_{j,i}\)
  to describe agents who cannot see the value of the bit at position
  \((j,i)\) of the matrix, in an agent logic that extends
  propositional logic with a propositional background theory, in this
  case hardwiring the above description of the scenario (every agent
  sees all bits except one per row, and every bit is seen by some
  agent).  The common knowledge resulting from the visibility
  conditions is hence
  \begin{equation*}\textstyle
    C_\top \big(\bigwedge_{ij} (p_{j,i} \implies C_{\neg h_{j,i}}p_{j,i}) \land (\neg p_{j,i} \implies C_{\neg h_{j,i}}\neg p_{j,i})\big).
  \end{equation*}
  We write \(\alpha_j^{\le x}\) for the (purely propositional) formula stating that at most~$x$ bits are set in row~$j$; that is, $\alpha_j^{\le x}$ is the disjunction of all conjunctions of the form $\Land_{i\in H}p_{i,j}\land\Land_{i\in\{1,\dots,k\}\setminus H}\neg p_{j,i}$ where $H\subseteq\{1,\dots,k\}$ and $|H|\le x$.
  (This formula is of exponential size in the number~$k$ of columns, but note that this happens already in the original muddy children puzzle, i.e.\ in the case $n=1$.)
  The initial knowledge available to the agents before the first round is that at least one bit is set in each row:
  \[\textstyle C_\top(\bigwedge_j \neg\alpha_j^{\le 0}).\]
  In each communication round the agents then choose one row, and communicate whether or not they know the value of the bit they cannot see in that row.
  Assuming all agents do not know the value of their respective bit, this establishes common knowledge about everyone's uncertainty:
  \[\textstyle C_\top(\bigwedge_{ij} \neg C_{h_{j,i}}p_{j,i}  \land \neg C_{h_{j,i}}\neg p_{j,i}).\]
  Of course the order of rounds is irrelevant here, and the state of the protocol can hence simply be represented by a tuple \((x_i,\dots,x_n)\) where each \(x_j\) counts how many communication rounds have taken place for row \(j\) (counting communication of the initial knowledge that at least one bit is set in each row). The key invariant of the protocol is that if these counters reach $(x_1,\dots,x_n)$ without anyone having learned new bits, then this results in the accumulated common knowledge
  \[\textstyle C_\top(\bigwedge_j \neg\alpha_j^{\leq x_j}).\]
  This clearly holds in the beginning of the game due to the initial
  knowledge.  Then in state \((x_1,\dots,x_n)\), after querying
  row~\(j\), the common knowledge increases according to the inference
  \begin{multline*}
     \Gamma, C_\top(\neg\alpha_j^{\leq x_j}), \textstyle C_\top(\bigwedge_{i} \neg C_{h_{j,i}}p_{j,i}  \land \neg C_{h_{j,i}}\neg p_{j,i}) \\\vDash C_\top(\neg\alpha_j^{\le x_j + 1}).
   \end{multline*}
   where~$\Gamma$ represents the visibility axioms and~$\vDash$ denotes local (i.e.\ per-world) consequence..
  The formal proof is similar to the textbook proof for the original puzzle~\cite{DBLP:books/daglib/0017977}, and sketched as follows:
  \begin{itemize}
  \item Assume \(\alpha_j^{\le x_j + 1}\).
  \item From the accumulated knowledge of the previous rounds, the agents already know that  more than \(x_j\) bits are set in row~$j$. From the assumption, they can conclude that  exactly \(x_j + 1\)  bits are set.
  \item Given that nobody knew whether their respective missing bit in row~$j$ is set, the agents can conclude that more than \(x_j + 1\) bits are set in row~$j$, contradicting the assumption. (Otherwise at least one agent whose bit is set would see only \(x_j\) many set bits and could hence have deduced that her missing bit is set.)
  \end{itemize}
  Similar reasoning is used to conclude missing bits once enough communication rounds have been performed.
\end{example}

\subsection{Upper Bound: Encoding into the Full $\mu$-Calculus }\label{sec:embedd-ciel-mualc}
\newcommand{\catrans}{t}
\newcommand*{\leadsfrom}[0]{\ensuremath{\mathrel{\reflectbox{\ensuremath{\leadsto}}}}}
\newcommand{\ALCIFEs}[2]{\ensuremath{#1 \leadsto #2}}
\newcommand{\ALCIFEops}[2]{\ensuremath{#2 \leadsfrom #1}}

\newcommand{\ALCIFE}[2]{\ensuremath{\progbox{\succ} (\progdiamond{\pi_1} #1 \implies \progbox{\pi_2} #2)}}
\newcommand{\ALCIFEop}[2]{\ensuremath{\progbox{\pi_2^-} (\progdiamond{\pi_1} #1 \implies \progbox{\succ^-} #2)}}
\renewcommand*{\succ}{\mathsf{edge}}

\noindent We establish the \ExpTime upper bound on satisfiability
checking by a satisfiability-preserving translation of \ac{CIEL} into
the $\mu$-calculus with converse, also known as the \emph{full
  $\mu$-calculus}, whose satisfiability problem is in
\ExpTime~\cite{Vardi98}. We emphasize that the full $\mu$-calculus
does not have the finite model property~\cite{Vardi98,Streett82}; we
therefore establish a bounded model property separately in the next
section.

In the translation, we use fixpoints to take transitive-reflexive
closures, and inverse roles to close under symmetry. The main idea is
then to view the family $\sim$ of per-agent indistinguishability
relations~$\sim_a$ featuring in the definition of \ac{CIEL} models as
a ternary relation on a single domain, and to encode this ternary
relation as a binary relation between worlds and (agent,
world)-pairs. 

In fact, the single-variable fragment of the full $\mu$-calculus
suffices for the translation; we briefly recall its syntax and
semantics. The syntax is parametrized over sets \(\AProps\) and
\(\Prog\) of \emph{atomic propositions} and \emph{atomic programs},
respectively. %
A \emph{program} is either an atomic program or a \emph{converse
  program}~$\alpha^-$ of $\alpha\in\Prog$. Moreover, we fix a single
\emph{fixpoint variable}~$\var$. Then, the set of
\emph{formulae}~$\phi,\psi,\dots$ is given by the grammar
\begin{equation*}
  \phi,\psi::=\bot\mid \atom\mid \var\mid  \neg \phi\mid \phi\land\psi \mid \progbox{\alpha}\phi\mid \nu\var.\,\phi
\end{equation*}
where $\atom\in\AProps$ and~$\alpha$ is a program. The box operator
$\progbox{\alpha}$ is read `for all
$\alpha$-successors'. %
The~$\nu\var$ operator takes greatest fixpoints, and binds~$\var$;
that is, an occurrence of~$\var$ is \emph{free} if it lies outside the
scope of any $\nu\var$.  Application of~$\nu\var$ is restricted to
formulae~$\phi$ in which every free occurrence of~$\var$ is
\emph{positive}, i.e.~lies under an even number of
negations~$\neg$. Further propositional connectives $\top$, $\lor$,
$\implies$, $\iff$ are defined as usual. Moreover, we define diamond
operators as
$\progdiamond{\alpha}\phi:=\neg\progbox{\alpha}\neg \phi$. (Also, one
can define least fixpoints~$\mu z.\,\phi$.)%

The semantics is defined over \emph{models}~$\M=(X,R,V)$ that consist
of a domain~$X$, an assignment~$R$ of a transition relation
$R(\alpha)\subseteq X\times X$ to every atomic
program~$\alpha\in\Prog$; and a valuation $V\colon\AProps\to\Pow(X)$
of the atomic
propositions.  %
The interpretation of converse programs is defined by
$R(\alpha^-)=\{(e,d)\mid (d,e)\in R(\alpha)\}$. The semantics of a
formula~$\phi$ is then given as a function
$\Sem{\phi}_\M\colon\Pow(X)\to\Pow(X)$ whose argument serves as the
extension of~$\var$, inductively defined by the expected clauses for
Boolean operators, $\Sem{\atom}_\M=V(\atom)$, and
\begin{align*}
  \Sem{z}_\M(U) & = U\\
  \Sem{\progbox{\alpha}\phi}_\M(U) & = \{w\in X\mid \forall v\in X.\,(w,v)\in R(\alpha)\\ & \qquad\qquad\qquad\qquad\qquad\implies v\in\Sem{\phi}_\M(U)\}\\
  \Sem{\nu z.\,\phi}_\M(U) & = \bigcup\{Z\subseteq X\mid Z\subseteq\Sem{\phi}_\M(Z)\}.
\end{align*}
One shows by induction that when $\nu z.\,\phi$ is well-formed, then
$\Sem{\phi}_\M$ is monotone w.r.t.~set inclusion, so
$\Sem{\nu z.\,\phi}_\M(U)$ as defined above is, by the Knaster-Tarski
fixpoint theorem, the greatest fixpoint of~$\Sem{\phi}_\M$.

In detail, the translation is then defined as follows.  We assume for
the translation that the agent logic is given as a fragment of the
full $\mu$-calculus. For clarity, we write~$u$ for the syntactic
embedding; we assume that each~$u(\psi)$ is \emph{closed}, i.e.~does
not contain a free occurrence of~$\var$.  As indicated previously, a
typical choice would be \ALC (modulo the usual slight syntactic
shifts), but for purposes of the complexity analysis, the agent logic
could in fact be the full $\mu$-calculus itself (whereas the
axiomatization introduced in the next section needs assumptions on the
agent logic that are not satisfied by the full $\mu$-calculus, such as
the finite model property).

\newcommand{\Ctrans}[2]{\nu\var.\, #2 \allowbreak \land
  (\ALCIFEs{#1}{\var}) \allowbreak \land (\ALCIFEops{#1}{\var})} We
use three fresh atomic programs \(\pi_1, \pi_2,\succ\). The intention
is that \(\succ\) relates worlds to (agent, world)-pairs, out of
which~\(\pi_1\) extracts the agent component, and \(\pi_2\) the world
component. Speaking more precisely, we do not insist that
\(\pi_1,\pi_2\) are functional, so when $(\world,r)\in R(\succ)$ in a
model, then~$r$ may in fact represent several (agent, world) pairs,
namely all pairs $(a,\oworld)$ such that $(r,a)\in R(\pi_1)$ and
$(r,\oworld)\in R(\pi_2)$ (and indeed it may happen that~$r$
represents no pair at all). As per Remark~\ref{rem:s5}, we do not need to
worry at this point about the fact that the induced
indistinguishability relations are not forced to be equivalence
relations.
As an intermediate step in the translation, we introduce forward and
backward binary \textquote{next-step} modalities by abbreviation:
\[\ALCIFEs{\psi}{\phi} \defeq \ALCIFE{\psi}{\phi},\]
is understood as \textquote{all worlds that are reached by a single
  forward indistinguishability step for an agent satisfying \(\psi\)
  satisfy~$\phi$}, and
\[\ALCIFEops{\psi}{\phi} \defeq \ALCIFEop{\psi}{\phi}\]
describes the opposite direction, \textquote{all worlds that are
  reached by a single backward indistinguishability step for an agent
  satisfying \(\psi\) satisfy~$\phi$}.  Using these two abbreviations to
model a step along the symmetrization of the indistinguishability
relation, we can encode the common knowledge modality, modelling
transitive-reflexive closure via a greatest fixpoint as indicated
above: We define the translation~$\catrans$ of a \ac{CIEL}
formula~$\phi$ into a formula~$\catrans(\phi)$ in the full $\mu$-calculus
recursively by
\begin{equation*}
  \catrans(C_\psi\phi)= \Ctrans{u(\psi)}{\catrans(\phi)}
\end{equation*}
and commutation with all other constructs. (Since no recursive calls
of~$\catrans$ are made on~\(\psi\), its duplication does not lead to
exponential blowup; recall that by the current assumption, the
translation~$u$ of agent formulae does essentially nothing.) We thus
obtain the following result.

\newcommand{\Rel}[0]{\ensuremath{R}}
\newcommand{\tp}[1]{\ensuremath{\Rel(\pi_{#1})}}
\newcommand{\tpop}[1]{\ensuremath{\Rel(\pi_{#1}^-)}}
\newcommand{\edg}[0]{\ensuremath{\Rel(\succ)}}
\newcommand{\edgop}[0]{\ensuremath{\Rel(\succ^-)}}
\begin{theoremEnd}[end]{theorem}[Upper complexity
  bound]\label{thm:ciel-etcomplete}
  If the agent logic is a fragment of the full $\mu$-calculus, then
  the satisfiability problem of \ac{CIEL} is in \ExpTime.
\end{theoremEnd}
\begin{proofEnd}
  We show that the translation $\catrans$ is really
  satisfiability-preserving, i.e.~that an \ac{CIEL} formula $\phi$ is
  satisfiable iff $\catrans(\phi)$ is satisfiable in the full
  $\mu$-calculus.

  \emph{\textquote{Only if}:} Given a \ac{CIEL} model
  \(\M = (\Worlds, \A, V_\Wl, \sim)\) we construct a \(\mu\)-calculus
  model
  \(\M' = (\Worlds' = \Agents \uplus \Worlds \uplus
  (\Agents\times\Worlds), \Rel, V)\) where we interpret the syntactic
  material (atomic programs and atomic propositions) mentioned in
  agent formulae in~$\phi$ over~$\Agents\subseteq \Worlds'$ like in
  the given agent model~$\A$, exploiting that the agent logic is a
  fragment of the full $\mu$-calculus. We assume w.l.o.g.~that the
  sets~$\Worlds$,~$\Agents$, and $\Agents\times\Worlds$ are already
  disjoint. We define the valuation \(V\) as extending the valuation
  induced by \(\A\) to interpret world atoms via \(V_\Wl\), and
  moreover put
  \begin{align*}
    \edg & =\{ (\world,(\agent,\oworld)) \mid \world \sim_\agent \oworld \text{ in } \M\}\\
    \tp1 &= \{((\agent,\world),\agent)\mid (\agent,\world)\in\Agents\times\Worlds\}\\
    \tp2 &= \{((\agent,\world),\world)\mid (\agent,\world)\in\Agents\times\Worlds\}.
  \end{align*}
  This model translation encodes the ternary relation into a
  heterogeneous world set where edges are encoded as pairs of agents
  and target states.  The agent and the target state are then
  extracted from a pair with~$\pi_1$, $\pi_2$ serving as projections.

  We show by induction on~$\phi$ that $\M,\world\models\phi$ iff
  $\M',\world\models\catrans(\phi)$, for $\world\in\Worlds$. Again,
  the cases for world atoms and Boolean operators are trivial; we do
  the case for the common knowledge operator $C_\psi$.

  We first note that due to symmetry of \(\sim\), we have that for 
  \(\world,\oworld,\agent \in \Worlds'\),
  \[(\world,(\agent,y)) \in \edg,\]
  (and then $((\agent,\oworld),\agent) \in \tp1$,
  $((\agent,\oworld),\oworld) \in \tp2$) exactly when
  \[((\agent,\world),\oworld)\in \edgop\] (and then
  $((\agent,\world), \agent) \in \tp1$,
  $(\world,(\agent,\world)) \in \tpop2$), namely, both hold iff
  $\world\sim_a \oworld$, equivalently $\oworld\sim_a \world$, in~$\M$.  It is hence clear
  that \(\ALCIFEs{\psi}{\chi}\) and \(\ALCIFEops{\psi}{\chi}\) are
  equivalent over the model~$\M'$. Over this model, we can thus
  equivalently modify the translation to
  \begin{equation*}
    \catrans(C_\psi\phi)=\nu\var.\, \phi \land
  (\ALCIFEs{\psi}{\var});
  \end{equation*}
  also, the above implies that \(\ALCIFEs{\psi}{(-)}\) acts as a box
  modality on \(\psi\)-successors (\(\ALCIFEs{\psi}{\chi}\) holds at a
  world iff~$\chi$ holds at all $\psi$-successors). %
  By Lemma~\ref{thm:C_trans}, the interpretation of $C_\psi$ in~$\M$ is a
  greatest fixpoint involving a box operator for the same relation
  $\bigcup_{\agent \in \SemMA{\psi}}\sim_\agent$. More formally, since~$\M$
  embeds into \(\M'\), and the associated inclusion map is a
  $p$-morphism~\cite{blackburn_rijke_venema_2001}, i.e.~a functional
  bisimulation, w.r.t.~these relations, it follows by the well-known
  fact that the (forward) $\mu$-calculus is bisimulation-invariant
  that satisfaction of $C_\psi\phi$ is preserved.

  \emph{\textquote{If}:} Given a \(\mu\)-calculus-model
  \(\M = (\Worlds, \Rel, V)\) we construct an \ac{CIEL} model
  \(\M' = (\Worlds, \A, V_\Wl, \sim)\).  We use the original
  set~$\Worlds$ of worlds also as the underlying set~$\Agents$ of the
  agent model~$\A$, whose structure we then obtain by just suitably
  restricting~$\M$. Similarly, we take $V_\Wl$ to be the restriction
  of~$V$ to world atoms, i.e.~we interpret world atoms in~$\M'$ like
  in~$\M$. We define the indistinguishability relation~$\sim_\agent$
  for $\agent\in\Agents=\Worlds$ as the symmetric closure of the
  relation
  \begin{multline*}
    \{ (\world, \oworld) \mid \exists (\world,\world') \in \edg.\\
    (\world',\agent) \in \tp1, (\world',\oworld) \in
    \tp2\}, %
  \end{multline*}
  exploiting that by Remark~\ref{rem:s5}, we do not actually need to care
  whether~$\sim_a$ is reflexive or transitive. We do enforce symmetry
  to ensure more straightforward agreement of the semantics on both
  sides.
  
  It is then clear that for each \(\agent \in \Agents\) and each agent
  formula~$\psi$, \( \SatAtA{\agent}{\psi}\) iff
  \(\agent\in\SemM{\psi}(U)\), noting that the latter does not depend
  on~$U$ since agent formulae are assumed to be closed.

  Now note that by the above construction of \(\sim\),
  \(\ALCIFEs{\psi}{(-)}\) and \(\ALCIFEops{\psi}{(-)}\) jointly act as
  a box modality on the \(\psi\)-successors according to~\(\sim\),
  where the symmetric closure is emulated by considering both
  \(\ALCIFEs{\psi}{(-)}\) and \(\ALCIFEops{\psi}{(-)}\).  Replacing
  the semantic clause for~$C_\psi$ by the fixpoint characterization as
  per Lemma~\ref{thm:C_trans}, we thus obtain that the truth conditions
  for~$C_\psi$ is the same as for its translation along~$\catrans$. As
  the clauses for Boolean operators agree as well, and the valuation
  of world atoms in~$\M'$ is just inherited from~$\M$, we obtain that
  for each $\world\in\Worlds$ and each world formula~$\chi$,
  $\M',\world\models\chi$ iff $\world\in\Sem{\catrans(\chi)}_\M(U)$
  (noting again that the latter set does not depend on~$U$ because
  $\catrans(\chi)$ is closed), so~$\M'$ serves as a witness for
  satisfiability of any formula~$\chi$ such that~$\catrans(\chi)$ is
  satisfied within~$\M$.
\end{proofEnd}
\begin{proof}[Proof sketch]
  We need to show that the translation~$\catrans$ is really
  satisfiability-equivalent. From a model \(\M,\world_0\) of an
  \ac{CIEL} formula~\(\phi\), with \(\M = (\Worlds, \A, V_\Wl,\sim)\),
  we construct a model $(\M',\world_0)$ of $\catrans(\phi)$, with
  \(\M'=(\Worlds',\Rel,V)\), as follows. Following the intention of
  the translation~$\catrans(\phi)$ as indicated above, we
  take~$\Worlds'$ to be the union of the set~$\Worlds$ of worlds, the
  set~$\Agents$ of agents (the underlying set of~$\A$), and the set
  $\Agents\times\Worlds$ of (agent, world) pairs, assuming
  w.l.o.g.~that these sets are disjoint. We interpret the syntactic
  material (atomic programs and atomic propositions) mentioned in
  agent formulae in~$\phi$ over~$\Agents\subseteq \Worlds'$ like in
  the given agent model~$\A$, exploiting that the agent logic is a
  fragment of the full $\mu$-calculus. Similarly, we interpret world
  atoms on~$\Worlds$ in the same way as in~$\M$. Finally, we
  interpret~$\pi_1,\pi_2$ as the expected projections
  $\tp1=\{((\agent,\world),\agent)\mid
  (\agent,\world)\in\Agents\times\Worlds\}$,
  $\tp2=\{((\agent,\world),\world)\mid
  (\agent,\world)\in\Agents\times\Worlds\}$, and $\succ$ as
  \begin{equation*}
    \edg=\{(\world,(\agent,\oworld))\mid \world\sim_\agent\oworld\text{ in }\M\}.
  \end{equation*}
  It is not hard to see that~$(\M',x)$ is really a model of~$\phi$.

  In the converse direction, given a model $(\M,\world_0)$ of the
  $\mu$-calculus formula $\catrans(\phi)$, with $\M=(\Worlds,\Rel,V)$,
  we construct a model $(\M',\world_0)$ of~$\phi$, with
  $\M'=(\Worlds,\A,V_\Wl,\sim)$, as follows. We use the original
  set~$\Worlds$ of worlds also as the underlying set~$\Agents$ of the
  agent model~$\A$, whose structure we then obtain by just suitably
  restricting~$\M$. Similarly, we take $V_\Wl$ to be the restriction
  of~$V$ to world atoms, i.e.~we interpret world atoms in~$\M'$ like
  in~$\M$. Finally, as indicated above, we define the
  indistinguishability relation~$\sim_\agent$ for
  $\agent\in\Agents=\Worlds$ as the symmetric closure of the relation
  \begin{multline*}
    \{(\world,\oworld)\mid \exists (\world,e)\in \edg.\\
    (e,\agent)\in \tp1,(e,\oworld)\in \tp2\}.
  \end{multline*}
  Again, it is not hard to check that~$(\M,x)$ is really a model
  of~$\phi$.
\end{proof}
\noindent In combination with Theorem~\ref{thm:ciel-hardness}, we thus have
\begin{corollary}[Complexity of \ac{CIEL}]
  If the agent logic is a fragment of the full $\mu$-calculus, then
  the satisfiability problem of \ac{CIEL} is \ExpTime-complete.
\end{corollary}
\begin{remark}\label{rem:nominals}
  Indeed the above encoding implies that one can raise the
  expressiveness of the agent logic to extensions of the full
  $\mu$-calculus that remain decidable in \ExpTime. One candidate is
  the \emph{full hybrid $\mu$-calculus}~\cite{SattlerVardi01}, which
  extends the full $\mu$-calculus with \emph{nominals},
  i.e.~propositional atoms denoting single objects. This opens the
  possibility of combining explicitly named agents in the standard
  sense with abstract groups of agents, as in the formula
  \begin{equation*}
    C_{\mathsf{John}\vee \exists\,\mathsf{hasFriend}.\,\!\mathsf{John}}\,\mathsf{Pub\_on\_Wednesdays},
  \end{equation*}
  which says that John and his friends know that their regular pub
  night is on Wednesdays.
\end{remark}
\newcommand*{\axsystem}{\ensuremath{C5}}
\section{Completeness and Bounded Models}\label{sec:axiomatization}

We axiomatize \ac{CIEL} in Hilbert style using the following
system~\(\axsystem\) of axioms and rules:
\begin{align*}
  &(T)&& C_\psi\phi \implies \phi\\
  & (\bot) &&\phi \implies C_\bot\phi\\ 
  & (K) &&C_\psi(\phi\implies\gamma) \implies (C_\psi \phi \implies C_\psi \gamma)\\
  & (4) &&C_\psi\phi \implies C_\psi C_\psi\phi\\
  & (5) &&\neg C_\psi\phi \implies C_\psi\neg C_\psi\phi\\
  &(\Ind) &&C_{\psi\lor\chi}(\phi \implies (C_{\psi}\phi\land C_\chi\phi)) \implies (\phi \implies C_{\psi\lor\chi}\phi)\\
  &(\nec)&&\frac{\phi}{C_\psi\phi}\qquad
  (\anti) \quad \frac{\gamma \implies \psi}{C_{\psi}\phi \implies C_{\gamma}\phi}
\end{align*}
Recall that by our running assumptions, the agent logic~\(\LA\) is
closed under all propositional connectives.  We write
$\axsystem\vdash\phi$ if an \ac{CIEL} formula~$\phi$ is derivable in
this system;~\(\phi\) is \emph{consistent} if
$\axsystem\not\vdash\neg\phi$. For a finite set~$\Gamma$ of formulae,
we write $\widehat\Gamma$ for the conjunction of all formulae
in~$\Gamma$, and we say that~$\Gamma$ is \emph{consistent}
if~$\widehat\Gamma$ is consistent. The system includes the usual~$S5$
axioms for common knowledge, reflecting normality (axiom~\((K)\)),
reflexivity (axiom~\((T)\)), transitivity (axiom~\((4)\)), and
Euclideanity (axiom~\((5)\)), as well as the usual necessitation
rule~\((\nec)\). As usual, the ($K$) axiom implies commutation of
$C_\psi$ with conjunction, and hence, together with the necessitation
rule ($\nec$), monotonicity and replacement of equivalents for
$C_\psi$.  Specific properties of~\ac{CIEL} are reflected in the
axiom~\((\bot)\), which, together with~\((T)\), says that~$C_\bot\phi$
holds almost vacuously, in the sense that it does not claim any agent
to know anything but requires~$\phi$ to be true in the current world;
in the rule~\((\anti)\), which says that $C_\psi\phi$ is antimonotone
in~\(\psi\), as requiring fewer agents to know~$\phi$ is a weaker
claim; and, centrally, in the induction axiom~\((\Ind)\), which
captures the fact that the indistinguishability
relation~$\sim_{A\cup B}$ for a union $A\cup B$ of two sets~$A,B$ of
agents is the reflexive-transitive closure of the union of the
indistinguishability relations~$\sim_A$,~$\sim_B$ for the original
sets. The rule \((\anti)\) implies replacement of equivalents in the
index of $C_\psi$ (from equivalence of $\psi$ and~$\psi'$, derive
equivalence of $C_\psi\phi$ and $C_{\psi'}\phi$). Via \((\anti)\), the
system depends on reasoning in the agent logic, which we will assume
to be completely axiomatized.

We note first that the induction axiom generalizes to multiple
disjuncts:
\begin{theoremEnd}[end]{lemma}\label{lem:ind-gen}
  Every formula of the form
  \begin{equation*}
    C_{\bigvee_{i=1}^n\psi_i}(\phi \implies
    (\textstyle\bigwedge_{i=1}^nC_{\psi_i}\phi)) \implies (\phi
    \implies C_{\bigvee_{i=1}^n\psi_i}\phi)
  \end{equation*}
  (for $n\ge 0$) is derivable in \(\axsystem\).
\end{theoremEnd}
\begin{proofEnd}
  Induction on~$n$. We keep use of monotonicity, replacement of
  equivalents, and propositional reasoning implicit. The case $n=0$ is
  by ($\nec$) and $(\bot)$, and the case $n=1$ by ($T$). For the
  inductive step $n\to n+1$, where $n\ge 1$, we note first that the
  inductive hypothesis implies derivability of the formula
  \begin{multline*}
    C_{\bigvee_{i=1}^{n+1}\psi_i}C_{\bigvee_{i=1}^{n}\psi_i}(\phi \implies
    (\textstyle\bigwedge_{i=1}^nC_{\psi_i}\phi)) \implies\\ C_{\bigvee_{i=1}^{n+1}\psi_i}(\phi
    \implies C_{\bigvee_{i=1}^n\psi_i}\phi)
  \end{multline*}
  using ($\nec$) (with $C_{\bigvee_{i=1}^{n+1}\psi_i}$) and ($K$). Using
  ($\anti$),and ($4$), we then derive
  \begin{equation*}
    C_{\bigvee_{i=1}^{n+1}\psi_i}(\phi \implies
    (\textstyle\bigwedge_{i=1}^nC_{\psi_i}\phi)) \implies C_{\bigvee_{i=1}^{n+1}\psi_i}(\phi
    \implies C_{\bigvee_{i=1}^n\psi_i}\phi)
  \end{equation*}
  and thus, conjoining both sides of the implication with
  $C_{\bigvee_{i=1}^{n+1}\psi_i}(\phi\to C_{\psi_{n+1}}\phi)$ and using
  commutation of $C_{\bigvee_{i=1}^{n+1}\psi_i}$ with conjunction,
  \begin{multline*}
  C_{\bigvee_{i=1}^{n+1}\psi_i}(\phi \implies
  (\textstyle\bigwedge_{i=1}^{n+1}C_{\psi_i}\phi)) \implies\\
  C_{\bigvee_{i=1}^{n+1}\psi_i}(\phi \implies
  (C_{\bigvee_{i=1}^n\psi_i}\phi\land C_{\psi_{n+1}}\phi)).
  \end{multline*}
  The inductive claim for $n+1$ then follows by chaining this
  implication with an instance of ($\Ind$).
\end{proofEnd}
\noindent We explicitly record soundness of the system:
\begin{theoremEnd}[end]{theorem}[Soundness]
  If \(\axsystem \vdash \phi\) then \(\models \phi\).
\end{theoremEnd}
\begin{proofEnd}
  Soundness of the axioms $(T)$, $(4)$, $(5)$, $(K)$ and the rule
  $(\nec)$ is standard for modal logics interpreted over equivalence
  relations. Soundness of~$(\bot)$ and the rule $(\anti)$ is
  clear. Soundness of the induction axiom
  \(C_{\psi \lor \chi}(\phi \implies (C_{\psi}\phi \land C_\chi\phi))
  \implies (\phi \implies C_{\psi \lor \chi}\phi)\) is seen as
  follows. Let \(\world\) a world in an \ac{CIEL}-model~$\M$ that
  satisfies
  \(C_{\psi \lor \chi }(\phi \implies (C_{\psi}\phi \land
  C_\chi\phi))\) and \(\phi\); we have to show that
  \(\SatAt{\world}{C_{\psi \lor \chi}\phi}\).  We need to show that
  \(\SatAt{\oworld}{\phi}\) for all worlds \(\oworld\) such that
  \(\world \sim_{\SemMA{\psi \lor \chi}} \oworld\).  Let \(\oworld\)
  be such a world.  Then there exists a finite chain
  \(\world = \world_0 \sim_{\agent_1} \world_1 \sim_{\agent_2} \dots
  \sim_{\agent_{m - 1}} \world_{m - 1} \sim_{\agent_m} \world_m =
  \oworld\) such that
  \(\agent_1,\dots,\agent_m \in \SemMA{\psi \lor \chi}\).  By
  straightforward induction on~$i$, every state~$x_i$ in this chain
  satisfies $C_\psi\phi$, $C_\chi\phi$, and~$\phi$. In
  particular, $x_m$ satisfies~$\phi$, as required.
\end{proofEnd}
\noindent We show completeness via a finite canonical model
construction that is related to the standard treatment of \ac{PDL} in
that it needs to close canonical models under
transitivity~\cite{FischerLadner79,blackburn_rijke_venema_2001}. This construction
requires some restrictions on the agent logic:
\newcommand{\aclosure}[0]{\ensuremath{\mathsf{Clo_\Ag}}}
\begin{definition}\label{def:ag-closure}
  We say that the agent logic~\(\LA\) has the \emph{filtered model
    property} if, for each finite set~$\Sigma_\Ag$ of agent formulae
  that is closed under subformulae, there is an agent
  model~$\A(\Sigma_\Ag)$, with underlying set denoted by~$\Agents(\Sigma_\Ag)$,
  such that on the one hand any two distinct agents
  in~$\Agents(\Sigma_\Ag)$ are distinguished by a formula
  in~$\Sigma_\Ag$ (in particular~$\Agents(\Sigma_\Ag)$ is finite,
  namely at most exponential in $|\Sigma_\Ag|$), and on the other hand
  for every satisfiable subset $\Gamma\subseteq\Sigma_\Ag$, there is
  an agent in~$\Agents(\Sigma_\Ag)$ that satisfies all formulae
  in~$\Gamma$.  Since the agent logic is closed under propositional
  connectives, we then have, for each agent
  \(\agent\in\Agents(\Sigma_\Ag)\), a \emph{characteristic
    agent formula}~\(\hat{\agent}\) (a propositional combination of
  $\Sigma_\Ag$-formulae) such that
  \(\SemMA[\A(\Sigma_\Ag)]{\hat{\agent}} = \{\agent\}\). We put
  \begin{equation*}
    \aclosure(\Sigma_\Ag)=\Sigma_\Ag\cup\{\hat\agent\mid\agent\in\Agents(\Sigma_\Ag)\}.
  \end{equation*}
\end{definition}
\begin{example}
  As indicated in the introduction, $\ALC$ has the filtered model
  property (and is completely axiomatized), and in fact one can go
  beyond~$\ALC$ to some degree. In particular, the extension of~$\ALC$
  with \emph{nominals}, $\ALCO$, still has the filtered model
  property. (See also comments in~Remark~\ref{rem:nominals}.)
\end{example}
\newcommand{\target}{\rho_0}
\noindent We \emph{fix from now on a consistent \ac{CIEL}
  formula~$\target$}. We base our canonical model construction on a
suitable notion of closure. 

\newsavebox{\snegs}
\savebox{\snegs}{\smash{\(\dot{\neg}\)}}
\newcommand{\sneg}[1]{\usebox{\snegs}#1}%

\begin{definition}[Normalized negation]
  We let \(\sneg\phi=\chi\) if~\(\phi\) has the form
  \(\phi = \neg\chi\), and \(\sneg\phi=\neg\phi\) otherwise.
\end{definition}
\begin{definition}[Closure]\label{def:closure} %
  Let \(\Sigma_\Ag\) be the closure of the set of agent formulae
  occurring in \(\target\) under taking subformulae. Then the
  \emph{closure} \(\Sigma\) of~$\target$ is the least set of world
  formulae containing~$\target$ that is closed under (world)
  subformulae and normalized negation, and moreover satisfies
  \begin{equation*}
    \text{if }C_\chi\phi \in \Sigma\text{ and } \psi \in \aclosure(\Sigma_\Ag)\text{, then }
    C_\psi\phi \in
    \Sigma
\end{equation*}
  (with \(\aclosure(\Sigma_\Ag)\) as per Definition~\ref{def:ag-closure}).
\end{definition}

\noindent We next construct a weak form of model of~$\target$ that
assigns indistinguishability relations to sets of agents without
regard to their definition via reflexive-transitive closure; this will be rectified in a subsequent
step.
\begin{definition}[Pseudo-model]
  An \ac{CIEL} \emph{pseudo-model} $\M^p=(\Worlds,\A,V_\Wl,\sim^p)$
  consists of a set~$\Worlds$ of worlds, an agent model~$\A$ with
  underlying set~$\Agents$ of agents, a valuation
  $V_\Wl\colon\Atoms\to\Pow(\Worlds)$ of the world atoms, and an
  equivalence relation~$\sim^p_A$ on~$\Worlds$ for each subset
  $A\subset\Agents$. The \emph{semantics} of~\ac{CIEL} over pseudo-models
  is defined like over models, %
  except
  that the interpretation of~$C_\psi$ uses the relation~$\sim^p_{\SemMA{\psi}}$ in
  place of~$\sim_{\SemMA{\psi}}$.
\end{definition}
\noindent We construct a \emph{canonical pseudo-model}
\(\C^p_\Sigma = (\Worlds_\Sigma,\allowbreak \A(\Sigma_\Ag),\allowbreak V_\Wl, \sim^p)\) by
taking~$\Worlds_\Sigma$ to consist of the maximal consistent subsets
of~\(\Sigma\); \(\A(\Sigma_\Ag)\) as per Definition~\ref{def:ag-closure};
$V_\Wl(p)=\{\Gamma\in\Worlds_\Sigma\mid p\in\Gamma\}$; and
\[\Gamma \sim^p_{\Sem{\psi}_\A} \Delta \iff \hat{\Gamma} \land P_\psi\hat{\Delta}
  \text{ consistent}\] where by~$P_\psi$ we denote the dual
of~$C_\psi$, i.e.~$P_\psi\phi:=\neg C_\psi\neg\psi$. Well-definedness
is guaranteed by the properties of~$\A(\Sigma_\Ag)$; indeed we shall
often write~$\sim^p_\psi$ in place of $\sim^p_{\Sem{\psi}_\A}$ for
readability, similarly for~$\sim$.
We note that~$\Sigma$ inherits exponential size
from~$\aclosure(\Sigma_\Ag)$, so~$\Worlds_\Sigma$ is of doubly
exponential size in the size of~$\target$.

\begin{theoremEnd}[end]{lemma}\label{thm:sim_refl_sym}
  The relations~$\sim^p_\psi$ are reflexive and symmetric.
\end{theoremEnd}
\begin{proofEnd}
  \emph{Symmetry:} Let \(\hat{\Gamma} \land P_\psi\hat{\Delta}\) be consistent.
  Then \(\hat{\Delta} \land P_\psi\hat{\Gamma}\) is consistent; for assume otherwise.
  Then \(\vdash \hat{\Delta} \implies C_\psi\neg\hat{\Gamma}\), and hence \(\hat{\Gamma} \land P_\psi C_\psi\neg\hat{\Gamma}\) is consistent.
  By the contraposition of axiom \((5)\), we obtain that \(\hat{\Gamma} \land C_\psi\neg\hat{\Gamma}\) is consistent, which contradicts \((T)\).

  \emph{Reflexivity:} To show that $\Gamma\sim^p_\psi\Gamma$, suppose otherwise.  Then
  \(\vdash \hat{\Gamma} \implies C_\psi\neg\hat{\Gamma}\) and via
  \((T)\), \(\vdash \hat{\Gamma} \implies \neg\hat{\Gamma}\),
  contradicting consistency of~$\Gamma$.
\end{proofEnd}

\noindent The point of using the canonical pseudo-model is that it allows for a straightforward proof of the usual existence lemma:
\begin{theoremEnd}[end]{lemma}[Existence Lemma]\label{thm:existence}
  Let \(\Gamma \in \Worlds_\Sigma\) be a world in the canonical
  pseudo-model~$\C^p_\Sigma$ such that \(\neg C_\psi\phi \in
  \Gamma\). Then there exists \(\Delta \in \Worlds_\Sigma\) such that
  \(\Gamma \sim^p_\psi \Delta\) and \(\neg\phi \in \Delta\)
\end{theoremEnd}
\begin{proofEnd}
  Enumerate the formulae
  \(\sigma_1,\dots,\sigma_m \in \Sigma\).  We define sets~\(\Delta_n\)
  iteratively to make \(\hat{\Gamma} \land P_\psi\hat{\Delta_n}\)
  consistent:
  \begin{description}
  \item[\(0\):] \(\Delta_0 = \{\neg\phi\}\). Since
    \(\neg C_\psi\phi \in \Gamma\), we trivially have that
    \(\hat{\Gamma} \land P_\psi\neg\phi\) is consistent.
  \item[\(n \to n+1\):] As \(\hat{\Gamma} \land P_\psi\hat{\Delta_n}\)
    is consistent, so is
    \(\hat{\Gamma} \land P_\psi(\hat{\Delta_n} \land (\sigma_{n+1}
    \lor \sneg\sigma_{n+1}))\). Since~$P_\psi$ derivably commutes with
    disjunction, it follows that
    \(\hat{\Gamma} \land P_\psi((\hat{\Delta_n} \land \sigma_{n+1})
    \lor (\hat{\Delta_n} \land \sneg\sigma_{n+1}))\) is consistent,
    i.e.\ at least one of
    \(\hat{\Gamma} \land P_\psi(\hat{\Delta_n \cup
      \{\sigma_{n+1}\}})\) or
    \(\hat{\Gamma} \land P_\psi(\hat{\Delta_n \cup
      \{\sneg\sigma_{n+1}\}})\) is consistent, and correspondingly
    $\Delta_n \cup \{\sigma_{n+1}\}$ or
    $\Delta_n \cup \{\sneg\sigma_{n+1}\}$ can then be picked as
    \(\Delta_{n+1}\).
  \end{description}
  Then, \(\Delta_m\) serves as the desired \(\Delta\).
\end{proofEnd}
\noindent Leveraging characteristic agent formulae, we can derive a
proper \ac{CIEL}-model, the \emph{canonical model}~$\C_\Sigma$, from
the canonical pseudo-model~$\C^p_\Sigma$ by taking
\[\sim_\agent\; =\; \sim^p_{\hat{\agent}},\]
where we exploit that by Remark~\ref{rem:s5}, we do not need to care
whether~$\sim_a$ is transitive. The key point is then that the
existence lemma survives this transition thanks to the following fact,
which hinges on the induction axiom~\((\Ind)\):
\begin{theoremEnd}[end]{lemma}\label{thm:pre-relation}
  For all formulae \(\psi \in \aclosure(\Sigma_\Ag)\), \(\sim^p_\psi \subseteq \sim_{\SemMA{\psi}}\)
\end{theoremEnd}
\begin{proofEnd}
  Let \(\Gamma, \Delta \in \Worlds_\Sigma\), and let
  \(\psi \in \aclosure(\Sigma_\Ag)\) be an agent formula such that
  \(\Gamma \sim^p_\psi \Delta\).  We have to show that there exist a
  finite sequence of worlds \(C_0,\dots,C_n\) and
  \(\agent_0,\dots,\agent_{n-1} \in \SemMA{\psi}\) such that
  \(\Gamma = C_0 \sim_{\agent_0} C_1, \dots , C_{n-1}
  \sim_{\agent_{n-1}} C_n = \Delta\).  Let \(D\) be the set of all
  worlds reachable from \(\Gamma\) by such a sequence; we thus have to
  show that \(\Delta \in D\).  Define
  \(\delta \defeq \bigvee_{\Theta \in D}\hat{\Theta}\). Note that
  \(\delta \land \bigvee_{\agent \in \SemMA{\psi}}
  P_{\hat{\agent}}\neg\delta\) is inconsistent, for suppose otherwise.
  Then (by derivable commutation of~$P$ operators with disjunction),
  \(\delta \land P_{\hat{\agent}}\hat{F}\) would be consistent for
  some \(\agent \in \SemMA{\psi}\) and some \(F_{\agent} \notin D\).
  Hence, also \(\hat{C} \land P_{\hat{\agent}}\hat{F_{\agent}}\) is
  consistent for at least one \(C \in D\).  But then
  \(C \sim^p_{\hat{\agent}} F_{\agent}\),
  i.e.~\(C \sim_{\agent} F_{\agent}\), implying that \(F_{\agent}\) is
  reachable from \(\Gamma\), in contradiction to
  \(F_{\agent} \notin D\).  So,
  \(\vdash \delta \implies \bigwedge_{\agent \in
    \SemMA{\psi}}C_{\hat{\agent}}\delta\).  By \((\nec)\), we obtain
  \(\vdash C_{\bigvee_{\agent \in \SemMA{\psi}}\hat{\agent}}(\delta
  \implies \bigwedge_{\agent \in
    \SemMA{\psi}}C_{\hat{\agent}}\delta)\), whence
  \(\vdash \delta \implies C_{\bigvee_{\agent \in
      \SemMA{\psi}}\hat{\agent}}\delta\) by Lemma~\ref{lem:ind-gen}.
  From Lemma~\ref{thm:sim_refl_sym}, \(\Gamma \in D\) and hence
  \(\vdash \hat{\Gamma} \implies C_{\bigvee_{\agent \in
      \SemMA{\psi}}\hat{\agent}}\delta\).  By the assumptions
  on~$\A(\Sigma_\Ag)$,
  \(\vdash \psi \implies \bigvee_{\agent \in
    \SemMA{\psi}}\hat{\agent}\).  Using \((\anti)\), we obtain
  \(\vdash \hat{\Gamma} \implies C_{\psi}\delta\).  As
  \(\Gamma \sim^p_\psi \Delta\),
  \(\hat{\Gamma} \land P_\psi\hat{\Delta}\) is consistent, and hence
  \(P_\psi(\hat{\Delta} \land \delta)\) is consistent by standard
  reasoning with the~$(K)$ axiom. This means that
  $\hat{\Delta} \land \delta$ is consistent, whence
  \(\hat{\Theta} \land \hat{\Delta}\) is consistent for at least one
  disjunct \(\hat{\Theta}\) of \(\delta\) (with~$\Theta\in D$).  As
  both \(\Theta\) and~\(\Delta\) are maximal consistent subsets of
  \(\Sigma\), we obtain \(\Theta = \Delta\) and hence
  \(\Delta \in D\).
\end{proofEnd}
\noindent It is then straightforward to establish the expected truth
lemma, making use of the fact that $\aclosure(\Sigma_\Ag)$ contains
all requisite characteristic agent-formulae:
\begin{theoremEnd}[end]{lemma}[Truth Lemma]\label{thm:truth}
  Let \(\Gamma\) be a world in the canonical model \(\C_\Sigma\). Then
  for all \(\phi \in \Sigma\),
  \[\phi \in \Gamma \iff \SatAt[\C_\Sigma]{\Gamma}{\phi}.\]
\end{theoremEnd}
\begin{proofEnd}
  Induction on \(\phi\).
  The propositional cases are trivial.
  In the case for \(\phi = C_\psi\chi\), we need to show that \(C_\psi\chi \in \Gamma\) iff \(\SatAt[\C_\Sigma]{\Gamma}{C_\psi\chi}\).

  \emph{`Only if':} Suppose that \(C_\psi\chi \in \Gamma\).  We need
  to show that \(\SatAt[\C_\Sigma]{\Gamma}{C_\psi\chi}\), i.e.\ that
  all worlds reachable from \(\Gamma\) in finitely many \(\psi\)-steps
  satisfy \(\chi\).  By the inductive hypothesis, this means that all
  these worlds contain~$\chi$, which follows by \((T)\) once we show
  that all these worlds contain~$C_\psi\chi$; to this end, it suffices
  so show that the formula~$C_\psi\chi$ is inherited from any world
  (say,~$\Gamma$) to all its $\psi$-successors.  Assume the contrary.
  Then there exists a world \(\Delta\) such that
  \(\hat{\Gamma} \land P_{\hat{\agent}}\hat{\Delta}\) is consistent
  for some \(\agent \in \SemMA{\psi}\) and
  \(\neg C_\psi\chi \in \Delta\).  In particular,
  \(\hat{\Gamma} \land P_{\hat{\agent}}{\neg C_\psi \chi}\) is
  consistent,
  i.e.~\(\hat{\Gamma} \land \neg C_{\hat{\agent}}C_\psi\chi\) is
  consistent.  By the assumptions on~$\A(\Sigma_\Ag)$, and because
  \(\agent \in \SemMA{\psi}\), we have
  \(\vdash \hat\agent \implies \psi\).  Using \((\anti)\), we obtain
  \(\vdash C_\psi C_\psi\chi \implies C_{\hat{\agent}}C_\psi\chi\),
  and hence
  \(\vdash\neg C_{\hat{\agent}}C_\psi\chi\implies \neg C_\psi
  C_\psi\chi \).  Thus, \(\hat{\Gamma} \land \neg C_\psi C_\psi\chi\)
  is consistent; by (the contraposition of)~$(4)$, it follows that
  \(\hat{\Gamma} \land \neg C_\psi\chi\) is consistent, in
  contradiction to~$C_\psi\chi\in\Gamma$. %

  \emph{`If':} By contraposition. Suppose that
  \(C_\psi\chi \notin \Gamma\), so \(\neg C_\psi\chi \in \Gamma\). We
  need to show that \(\C_\Sigma,\Gamma\not\models C_\psi\chi\).
  By Lemma~\ref{thm:existence}, we have $\Gamma\sim^p_\psi\Delta$ for
  some~$\Delta$ such that $\neg\chi\in\Delta$,
  i.e.~$\chi\notin\Delta$. By Lemma~\ref{thm:pre-relation},
  $\Gamma\sim_{\SemMA{\psi}}\Delta$, and by the inductive hypothesis,
  \(\C_\Sigma,\Delta\not\models \chi\), showing the
  claim.%
  
\end{proofEnd}
\noindent Completeness and the bounded model property are then immediate:
\begin{theoremEnd}[end]{corollary}[Completeness over finite models]
  Suppose that \(\LA\) has the filtered model property and is
  completely axiomatized. Then \(\axsystem\), together with the
  axiomatization of~\(\LA\), is weakly complete, i.e.~every valid
  \ac{CIEL} formula is derivable. Moreover, \ac{CIEL} has the
  \emph{bounded model property}: if a formula~$\target$ is
  satisfiable, then~$\target$ has a finite model with at most doubly
  exponentially many worlds in the size of~$\target$.
\end{theoremEnd}
\begin{proofEnd}
  By contraposition: Let \(\target\) a formula such that
  \(\axsystem \not\vdash \target\), then \(\neg \target\) is
  consistent with \(\axsystem\).  Let \(\Sigma\) be the closure of
  \(\target\) as in Definition~\ref{def:closure}.  Then, there is a
  world \(\Gamma \in \Worlds_\Sigma\) in the canonical model
  \(\C_\Sigma\) such that \(\neg \target \in \Gamma\) (since in the
  finite ordered set of consistent subsets of~$\Sigma$, every element
  is below a maximal one).  By Lemma~\ref{thm:truth},
  \(\SatAt[\C_\Sigma]{\Gamma}{\neg\target}\).
\end{proofEnd}

\section{Conclusions}
We have introduced \emph{\acf{CIEL}}, a logic for reasoning about the
common knowledge of groups of agents that are described abstractly via
defining properties. We have established $\ExpTime$-completeness, a
bounded model property, and (necessarily weak) completeness of a
natural axiomatization. The $\ExpTime$ upper bound holds in spite of
the fact that the expected encoding into standard group epistemic
logic (with a common knowledge operator for enumerated groups of
agents) incurs exponential blowup, and relies instead on a
satisfiability-preserving translation into the $\mu$-calculus with
converse. Key directions for future research concern in particular
extensions of the logic by a distributed knowledge operator; by
dynamic epistemic modalities such as public announcements; by
expressive means for describing groups of agents via their individual
knowledge; and by allowing non-rigid agent names and agent atoms to
capture knowledge about agents.

\section{Acknowledgements}
This work was supported by DFG (German Research Foundation) as 
part of the Research Training Group~2475 ``Cybercrime and Forensic Computing'' 
(grant number~393541319/GRK2475/1-2019) and by DFG project ``RAND: Reconstructing Arguments from Newsworthy Debates'' (grant number~377333057).
The authors also wish to thank the anonymous reviewers for their suggestions and feedback.

\bibliography{ciel.bib}

\providecommand{\noopsort}[1]{}
\begin{thebibliography}{21}
\providecommand{\natexlab}[1]{#1}

\bibitem[{Baader et~al.(2003)Baader, Calvanese, McGuinness, Nardi, and
  Patel-Schneider}]{BaaderEA03}
Baader, F.; Calvanese, D.; McGuinness, D.; Nardi, D.; and Patel-Schneider, P.,
  eds. 2003.
\newblock \emph{The Description Logic Handbook}.
\newblock Cambridge University Press.
\newblock ISBN 0-521-78176-0.

\bibitem[{Baltag, Moss, and Solecki(1998)}]{BaltagEA98}
Baltag, A.; Moss, L.~S.; and Solecki, S. 1998.
\newblock The Logic of Public Announcements and Common Knowledge and Private
  Suspicions.
\newblock In Gilboa, I., ed., \emph{Theoretical Aspects of Rationality and
  Knowledge, TARK-1998}, 43--56. Morgan Kaufmann.

\bibitem[{B{\'{\i}}lkov{\'{a}}, Christoff, and
  Roy(2021)}]{DBLP:journals/corr/abs-2106-11493}
B{\'{\i}}lkov{\'{a}}, M.; Christoff, Z.; and Roy, O. 2021.
\newblock Revisiting Epistemic Logic with Names.
\newblock In \emph{Theoretical Aspects of Rationality and Knowledge, {TARK}
  2021}, volume 335 of \emph{{EPTCS}}, 39--54.

\bibitem[{Blackburn, de~Rijke, and Venema(2001)}]{blackburn_rijke_venema_2001}
Blackburn, P.; de~Rijke, M.; and Venema, Y. 2001.
\newblock \emph{Modal Logic}.
\newblock Cambridge Tracts in Theoretical Computer Science. Cambridge
  University Press.

\bibitem[{de~Rijke(1993)}]{derijkephd}
de~Rijke, M. 1993.
\newblock \emph{Extending Modal Logic}.
\newblock Ph.D. thesis, Universiteit van Amsterdam Institute for Logic,
  Language and Computation.

\bibitem[{Fagin et~al.(1995)Fagin, Halpern, Moses, and
  Vardi}]{DBLP:books/mit/FHMV1995}
Fagin, R.; Halpern, J.~Y.; Moses, Y.; and Vardi, M.~Y. 1995.
\newblock \emph{Reasoning About Knowledge}.
\newblock {MIT} Press.
\newblock ISBN 9780262562003.

\bibitem[{Fischer and Ladner(1979)}]{FischerLadner79}
Fischer, M.~J.; and Ladner, R.~E. 1979.
\newblock Propositional Dynamic Logic of Regular Programs.
\newblock \emph{J.\ Comput.\ Syst.\ Sci.}, 18(2): 194--211.

\bibitem[{Fitting, Thalmann, and
  Voronkov(2001)}]{DBLP:journals/sLogica/FittingTV01}
Fitting, M.; Thalmann, L.; and Voronkov, A. 2001.
\newblock Term-Modal Logics.
\newblock \emph{Stud Logica}, 69(1): 133--169.

\bibitem[{Grove(1995)}]{DBLP:journals/ai/Grovez95}
Grove, A.~J. 1995.
\newblock Naming and Identity in Epistemic Logic Part {II:} {A} First-Order
  Logic for Naming.
\newblock \emph{Artif. Intell.}, 74(2): 311--350.

\bibitem[{Grove and Halpern(1993)}]{DBLP:journals/logcom/GroveH93}
Grove, A.~J.; and Halpern, J.~Y. 1993.
\newblock Naming and Identity in Epistemic Logics Part {I:} The Propositional
  Case.
\newblock \emph{J. Log. Comput.}, 3(4): 345--378.

\bibitem[{Huth and Ryan(2004)}]{DBLP:books/daglib/0017977}
Huth, M.; and Ryan, M.~D. 2004.
\newblock \emph{Logic in Computer Science - Modelling and Reasoning about
  Systems {(2.} ed.)}.
\newblock Cambridge University Press.
\newblock ISBN 9780511264016.

\bibitem[{Lutz(2006)}]{Lutz06}
Lutz, C. 2006.
\newblock Complexity and succinctness of public announcement logic.
\newblock In \emph{Autonomous Agents and Multiagent Systems, AAMAS 2006},
  137--143. {ACM}.

\bibitem[{Mares(2020)}]{sep-logic-relevance}
Mares, E. 2020.
\newblock {Relevance Logic}.
\newblock In \emph{The {Stanford} Encyclopedia of Philosophy}. Metaphysics
  Research Lab, Stanford University, {W}inter 2020 edition.

\bibitem[{Moses and Tuttle(1988)}]{MosesTuttle88}
Moses, Y.; and Tuttle, M. 1988.
\newblock Programming Simultaneous Actions Using Common Knowledge.
\newblock \emph{Algorithmica}, 3: 121--169.

\bibitem[{Naumov and Tao(2019)}]{DBLP:journals/rsl/NaumovT19}
Naumov, P.; and Tao, J. 2019.
\newblock Everyone Knows that someone Knows: Quantifiers over Epistemic Agents.
\newblock \emph{Rev. Symb. Log.}, 12(2): 255--270.

\bibitem[{Padmanabha and Ramanujam(2019)}]{PadmanabhaRamanujam19}
Padmanabha, A.; and Ramanujam, R. 2019.
\newblock Two variable fragment of Term Modal Logic.
\newblock In \emph{Mathematical Foundations of Computer Science, {MFCS} 2019},
  volume 138 of \emph{LIPIcs}, 30:1--30:14. Schloss Dagstuhl -- Leibniz-Zentrum
  f{\"{u}}r Informatik.

\bibitem[{Pavlovic(2021)}]{Pavlovic21}
Pavlovic, D. 2021.
\newblock Probabilistic Annotations for Protocol Models.
\newblock In Dougherty, D.; Meseguer, J.; M{\"{o}}dersheim, S.; and Rowe, P.,
  eds., \emph{Protocols, Strands, and Logic}, volume 13066 of \emph{LNCS},
  332--347. Springer.

\bibitem[{Sattler and Vardi(2001)}]{SattlerVardi01}
Sattler, U.; and Vardi, M. 2001.
\newblock The Hybrid {\(\mathrm{\mu}\)}-Calculus.
\newblock In \emph{Automated Reasoning, {IJCAR} 2001}, volume 2083 of
  \emph{LNCS}, 76--91. Springer.

\bibitem[{Streett(1982)}]{Streett82}
Streett, R. 1982.
\newblock Propositional Dynamic Logic of Looping and Converse Is Elementarily
  Decidable.
\newblock \emph{Inf.\ Control.}, 54(1/2): 121--141.

\bibitem[{Vardi(1998)}]{Vardi98}
Vardi, M. 1998.
\newblock Reasoning about The Past with Two-Way Automata.
\newblock In \emph{Automata, Languages and Programming, ICALP 1998}, volume
  1443 of \emph{LNCS}, 628--641. Springer.

\bibitem[{Venema(1996)}]{venemaCrashArrow}
Venema, Y. 1996.
\newblock A Crash Course in Arrow Logic.
\newblock In \emph{Arrow Logic and Multi-Modal Logic}, Studies in Logic,
  Language, and Information. European Association of Logic, Language and
  Information (FoLLI).
\newblock ISBN 9781575869858.

\end{thebibliography}
\clearpage %
\appendix
\section{Details and Full Proofs}

We give details and full proofs where these have been omitted in the
main body of the paper; for the convenience of the reader, we restate
the relevant claims. When discussing \ac{CIEL} models, we use the
following additional terminology: When \(\world \sim_\agent \oworld\)
and \(\SatAtA{\agent}{\psi}\), we call \(\oworld\) a
\emph{\(\psi\)-successor of \(\world\)}.

\printProofs

\end{document}